\documentclass{article}
\usepackage{openwork}
\usepackage{amsmath, amssymb, amsfonts}
\usepackage{graphicx}
\usepackage{geometry}
\usepackage{booktabs}
\usepackage{tikz}
\usetikzlibrary{arrows.meta}
\usepackage{hyperref}
\usepackage{algorithm}
\usepackage{algorithmic}
\usepackage{subcaption}
\usepackage{float}
\usepackage{bm}
\usepackage{multirow}
\usepackage{xcolor}

\title{An Operator-Consistent Graph Neural Network for Learning Diffusion Dynamics on Irregular Meshes}
\author[1]{Yuelian Li}
\author[2]{Andrew Hands}
\affil[1]{Computational and Applied Mathematics, University of Chicago}
\affil[2]{Computer Science, University of Chicago}
\date{}

\begin{document}
\maketitle

\thispagestyle{plain}
\newpage
\tableofcontents
\newpage


\begin{abstract}
Classical numerical methods solve partial differential equations (PDEs) efficiently on regular meshes, but many of them become unstable on irregular domains. In practice, multiphysics interactions such as diffusion, damage, and healing often take place on irregular meshes. We develop an operator-consistent graph neural network (OCGNN-PINN) that approximates PDE evolution under physics-informed constraints. It couples node-edge message passing with a consistency loss enforcing the gradient-divergence relation through the graph incidence matrix, ensuring that discrete node and edge dynamics remain structurally coupled during temporal rollout. We evaluate the model on diffusion processes over physically driven evolving meshes and real-world scanned surfaces. The results show improved temporal stability and prediction accuracy compared with graph convolutional and multilayer perceptron baselines, approaching the performance of Crank-Nicolson solvers on unstructured domains.
\end{abstract}


\section{Introduction}
\label{sec:introduction}
Partial differential equations (PDEs) are a fundamental tool for describing spatiotemporal processes in physics, biology, and engineering. Classical numerical solvers such as finite difference (FD), finite element (FEM), and Crank–Nicolson (CN) schemes have long been used to approximate PDE solutions on structured meshes. However, these methods depend on regular discretizations, which restrict their use when the domain is irregular, evolving, or high-dimensional. Many physical systems---such as healing and damage in materials, or transport on curved biological surfaces---require solvers that remain stable and accurate on unstructured meshes.

Graph neural networks (GNNs) provide a natural way to represent and learn over irregular domains. Their message-passing operators can approximate discrete differential operators, such as gradients and Laplacians, while directly incorporating mesh connectivity. Compared with multilayer perceptron (MLP)-based physics-informed neural networks (PINNs), which rely on continuous coordinate encodings, GNN architectures preserve local geometric structure and handle arbitrary boundaries more flexibly. This makes them an appealing direction for physics-informed PDE learning on irregular meshes.

In this work, we propose \textbf{OCGNN-PINN}, an operator-consistent graph neural network model for physics-informed PDE learning on irregular meshes. The model combines geometry-aware message passing with energy-consistent regularization to align learned dynamics with the discrete structure of diffusion equations. It captures spatial anisotropy and local geometric relations through a hierarchical node-to-edge and edge-to-node message passing process, while the P-tensor consistency loss enforces gradient--divergence duality and energy conservation across node and edge dynamics. Coupled with explicit Laplacian constraints in the physics-informed loss, OCGNN-PINN\footnote{ Note that the term ``OCGNN'' here refers to operator consistency in graph-based PDE learning and is unrelated to the ``One-Class GNN'' framework previously proposed for anomaly detection.} connects learned graph propagation with the underlying physical operator. We evaluate the model on diffusion processes defined over irregular and physically driven evolving meshes that represent healing--damage dynamics and real-world scanned surfaces. Experimental comparisons demonstrate that the proposed approach achieves superior performance relative to standard graph convolutional and multilayer perceptron architectures, with solution quality comparable to established Crank-Nicolson numerical methods on unstructured geometries.

\section{Background}  

\subsection{Mesh Representations and Numerical Instability}

In numerical PDE solvers, the spatial domain is typically discretized into a computational mesh. 
When the mesh is uniformly spaced, classical finite difference (FD) or Crank--Nicolson (CN) schemes 
provide stable and accurate approximations of the Laplacian operator. 
However, when the mesh becomes irregular, the local finite difference stencil loses its symmetry, 
and numerical stability can no longer be guaranteed.  

To demonstrate this effect, we implemented a simple two-dimensional diffusion equation 
based on a standard finite difference implementation of the 2D diffusion equation. \footnote{The implementation is adapted from the SciPython demo\url{https://scipython.com/books/book2/chapter-7-matplotlib/examples/the-two-dimensional-diffusion-equation/}}. 
The uniform case uses a regular $100 \times 100$ grid with a time step chosen according to 
the Courant--Friedrichs--Lewy (CFL) stability condition \cite{CFL_source_citation}, 
which ensures that the explicit update remains numerically stable for uniform spacing. 
Here, $\Delta x$ and $\Delta y$ denote the spatial grid spacings in the $x$- and $y$-directions, 
$D$ is the diffusion coefficient, and $\Delta t$ is the time step. 
The stability condition is given by
\begin{equation}
    \Delta t \leq \frac{\Delta x^2 \Delta y^2}{2D(\Delta x^2 + \Delta y^2)}.
\end{equation}

This inequality ensures that information does not propagate faster than the numerical scheme can resolve, 
maintaining stability of the explicit finite-difference update.

To construct an irregular mesh, each grid coordinate is perturbed by a random offset 
of up to 60\% of the local grid spacing and clipped to the domain boundaries. 
Although the mean grid resolution and diffusion coefficient are kept identical, 
the nonuniform spacing alters the discrete Laplacian, breaking its isotropy and conservation properties.  

\begin{figure}[htbp]
    \centering
    \includegraphics[width=0.7\textwidth]{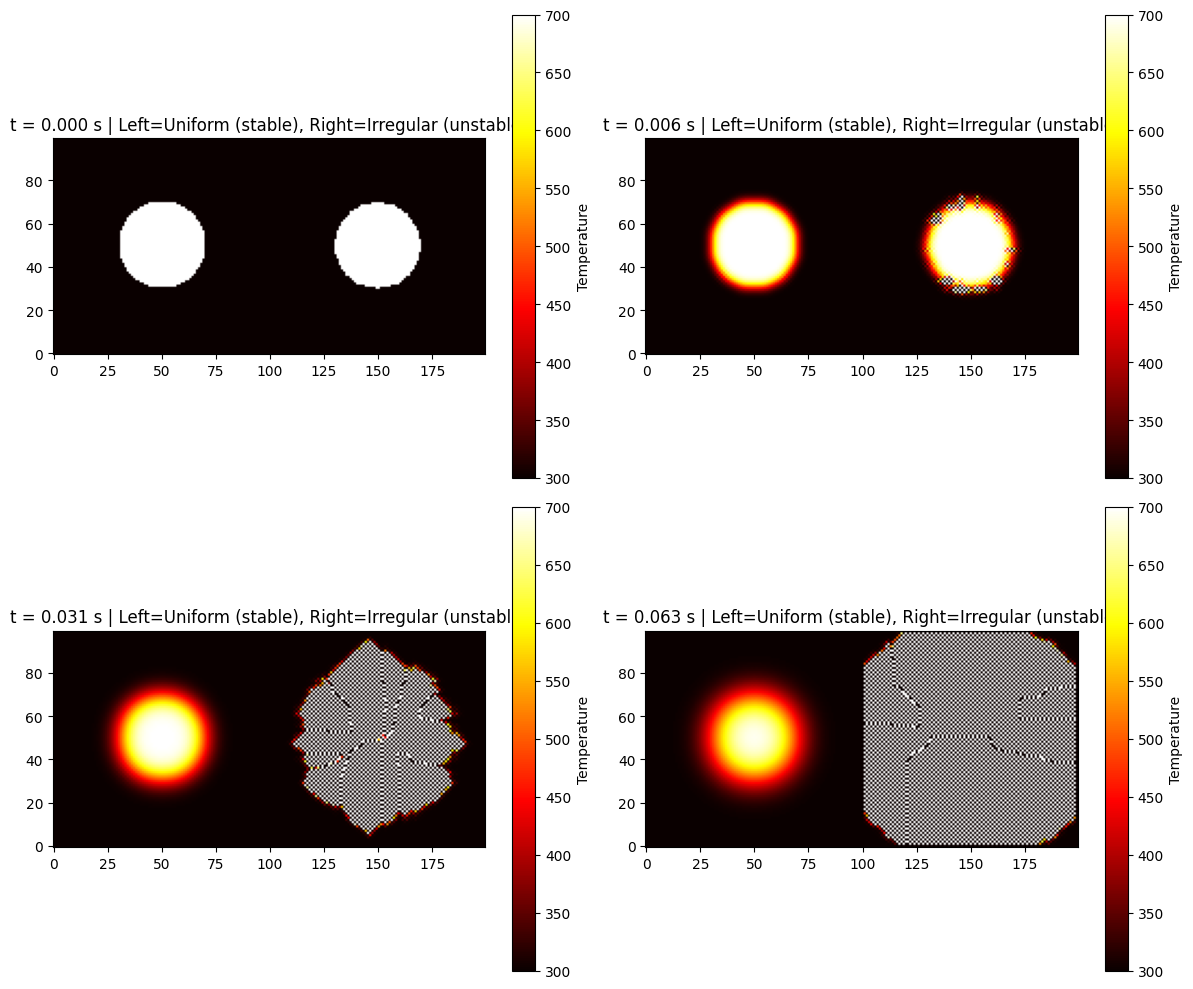}
    \caption{Temperature evolution in the 2D diffusion equation. 
    The uniform mesh (left) remains stable, while the irregular mesh (right) diverges under identical diffusion coefficients and time steps.}
    \label{fig:mesh_instability}
\end{figure}

Figure~\ref{fig:mesh_instability} compares the temperature evolution on the uniform and irregular grids 
under the same initial and boundary conditions. 
While the uniform mesh produces a stable, smooth diffusion profile, 
the irregular mesh exhibits spurious oscillations and eventually diverges, 
as the discretized Laplacian ceases to be negative definite. 
This simple example\footnote{See GitHub: \url{https://github.com/abyssyli/2025Summer/blob/main/PINNs/2D_Diffusion_Equation_on_Irregular_Mesh.ipynb}.} shows that even minor geometric perturbations 
can lead to catastrophic instability in classical solvers, 
underscoring the need for PDE learning methods that remain stable 
on unstructured or evolving meshes.

\subsection{From PINN to GNN: Bridging Continuous and Discrete Physics}

\subsubsection{Physics-Informed Neural Networks}

Physics-informed neural networks (PINNs)\cite{raissi2019physics} learn the solutions of partial differential equations (PDEs) by minimizing the residual of the governing equation together with boundary and initial conditions. 

For example, when applying PINN to solve the one-dimensional Heat Equation~\cite{crank1975diffusion},
\begin{equation}
\frac{\partial u}{\partial t} = \alpha \frac{\partial^2 u}{\partial x^2},
\end{equation}
the model takes the spatial position $x$ and time $t$ as inputs and outputs the predicted temperature $u(x, t)$. 

The loss function penalizes deviations from the PDE residual and the boundary conditions, driving the network toward the analytical solution.

Although PINNs do not require data supervision~\cite{raissi2019physics}, they rely on sampling collocation points from the continuous space-time domain to evaluate the PDE residuals~\cite{sundar2025sequential} and explicit formulations of the underlying equations. 
When the geometry becomes complex or analytical supervision is unavailable, this dependency limits their scalability and flexibility.

\subsubsection{Discretization and the Bridge to Graphs}

When PDEs are solved numerically using finite-difference or finite-element methods, the spatial domain is discretized into a set of nodes and elements. 

Nodes represent discrete spatial positions or physical quantities, while edges capture local interactions between adjacent nodes. 

Such discretized structures can naturally be represented as graphs, where the incidence relations~\cite{trask2022ddec} between nodes and edges correspond to numerical differential operators such as gradients and Laplacians.

This observation establishes a direct bridge between PDE solvers and graph-based learning. 
The same local connectivity that underlies finite-element discretizations can serve as the computational graph for a neural network designed to learn and propagate physical dynamics.

\subsubsection{From Residual Learning to Structural Learning}

Traditional PINNs focus on minimizing the PDE residual at sampled points, emphasizing the numerical accuracy of the solution. 

In contrast, GNNs learn local propagation rules that approximate the behavior of differential operators themselves, mapping node states to fluxes or higher-order interactions. 

This shift moves PDE learning from residual-based supervision to structure-based modeling, where the network encodes the relationships between physical operators such as gradient and divergence through its topology.

By explicitly encoding subgraphs or motifs, permutation equivariance\cite{{kondor2018generalization}} and physical symmetries can be incorporated directly into the network design. From this perspective, GNNs act not merely as data-driven approximators but as structure-preserving frameworks for representing PDE dynamics. 

When spatial and temporal relations can be reasonably inferred, GNNs become a more plausible and interpretable alternative to traditional PINNs.

\subsection{Necessity of Graph-based beyond interpolation and fixed Laplacians}

Although traditional finite-difference schemes can be extended to irregular domains 
through interpolation or local reconstruction, 
their discrete operators still depend on fixed stencils and predefined mesh topology. 
Once the discretization pattern is chosen, 
the notion of spatial coupling is effectively hard-coded, 
limiting adaptability to varying geometries or boundary conditions~\cite{leveque2007finite}.

Standard graph convolutional networks (GCNs) 
are formally equivalent to Laplacian-based diffusion processes, 
propagating features across edges through repeated neighborhood averaging~\cite{kipf2017semi}. 
While this structure captures smoothness on graphs, 
it remains fundamentally linear and static, 
lacking the capacity to model nonlinear dynamics, 
temporal evolution, or higher-order spatial couplings~\cite{oono2020graph}. 

Message-passing graph neural networks (MPNNs)
extend this framework by replacing the fixed finite-difference stencil 
with a learnable operator. 
Each edge-to-node interaction defines a localized update rule 
that can adaptively approximate differential relationships 
between variables on irregular domains~\cite{battaglia2018relational}. Building on this principle, 
Pfaff et~al.~(2021) demonstrated that learned message passing 
can effectively reproduce mesh-based physical simulations~\cite{pfaff2021learning}.
Rather than relying on a predetermined Laplacian, 
the network learns how information should propagate 
to best represent the underlying physical process~\cite{brandstetter2022message}. 

Therefore, even in the presence of interpolation-based finite-difference schemes 
or Laplacian-based graph convolutions, 
graph-based message passing remains valuable. 
It motivates the construction of our model, 
allowing neural networks to approximate the dynamics of partial differential equations 
on irregular domains in a physically informed manner.

\section{Irregular Mesh}
\label{sec:irregular_mesh}
\subsection{Construction of Irregular Mesh}
\subsubsection{Irregular Mesh on Ellipsoidal Domain}
The construction of irregular meshes in this study began with the manually designed samples used in our earlier simulations\footnote{See GitHub: \url{https://github.com/abyssyli/2025Summer/blob/main/PINNs/More_simple_irregular_mesh_samples_for_PINN_GNN_Model_for_2D_Difussion_Equation.ipynb}.}. 
A common practice for generating irregular meshes is to start from a uniform grid and introduce random perturbations with controllable amplitudes~\cite{bridson2007fast}, 
leading to jittered, clustered, or Poisson-type point distributions. 
While such meshes are easy to implement and can effectively test the robustness of graph-based solvers, 
they remain artificially created and lack any inherent physical or geometric basis. 

After confirming that finite-difference (FD) schemes become unreliable on these perturbed meshes 
whereas GNNs remain stable, 
we gradually shifted toward constructing a non-artificial, naturally merged, and physically driven irregular mesh. 

\begin{figure}[htbp]
    \centering
    \begin{subfigure}[t]{0.49\linewidth}
        \centering
        \includegraphics[width=\linewidth]{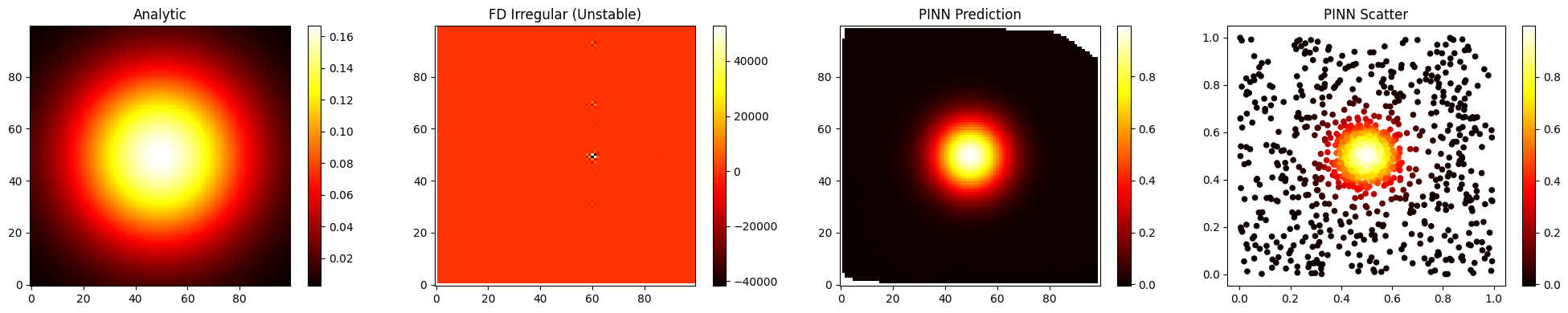}
        \caption{Comparison between analytic and GNN\_PINN solutions on a clustered irregular mesh.}
        \label{fig:clustered_mesh}
    \end{subfigure}
    \hfill
    \begin{subfigure}[t]{0.49\linewidth}
        \centering
        \includegraphics[width=\linewidth]{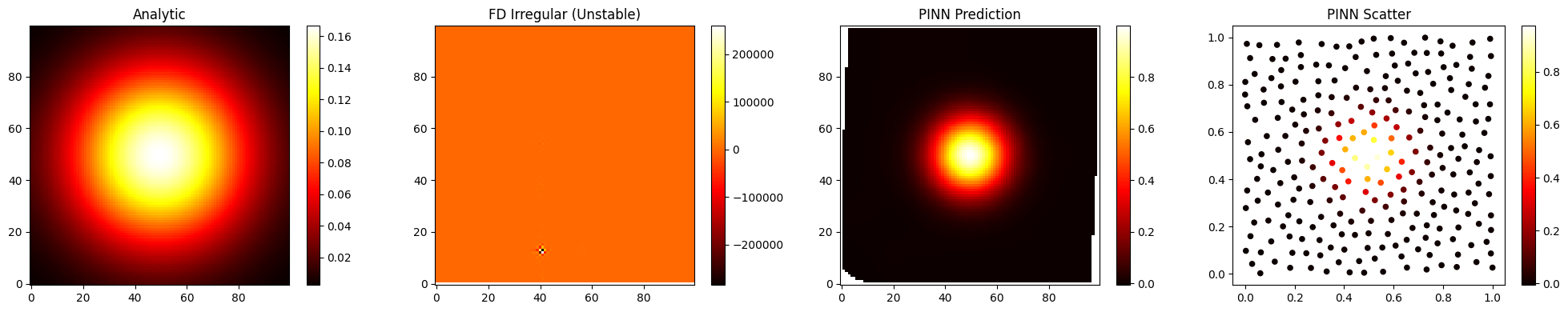}
        \caption{Comparison between analytic and GNN\_PINN solutions on a Poisson-distributed irregular mesh.}
        \label{fig:poisson_mesh}
    \end{subfigure}
    \caption{GNN\_PINN versus analytic comparison on clustered and Poisson irregular meshes. 
    Both geometries demonstrate that graph-based solvers remain numerically stable on irregular domains.}
    \label{fig:gnn_pinn_mesh_comparison}
\end{figure}

At the same time, considering how GNNs propagate information through message passing, 
the experimental mesh should not be a volumetric grid but rather a two-dimensional surface embedded in three-dimensional space. 
This motivates the development of a physically driven mesh generation process, 
which is described in detail in the following section.

\subsubsection{PDE Dynamics on Physically Driven Mesh}

We aim to construct an irregular mesh that arises from a physical process rather than random perturbations. 
In many natural systems, such as the branching of rivers, venation patterns on leaves, or biological tissue organization, 
geometric irregularity develops through the evolution of physical laws. 
In mathematical terms, such processes are often described by PDEs 
that determine how local interactions generate global structure. 
A physically driven irregular mesh should therefore reflect both geometric variability and physical consistency. 

To achieve this, we employ a coupled PDE system that governs the evolution of healing, damage\cite{enakoutsa2017physics}, and stress on a curved surface. 
This system is used to generate a surface geometry whose irregularity emerges from interpretable dynamics rather than arbitrary noise.

\paragraph{(1) Healing dynamics:}
\begin{equation}
\frac{\partial h}{\partial t}
= D_h \Delta h + \eta (1-h) - \lambda D.
\label{eq:healing}
\end{equation}
The variable $h(x,t)$ represents the local healing level on the surface. 
The diffusion term $D_h \Delta h$ promotes spatial spread of healing across neighboring regions, 
$\eta(1-h)$ drives the recovery toward a healed state, 
and the inhibitory term $-\lambda D$ suppresses healing in areas with high damage $D(x,t)$. 
These competing effects allow local recovery fronts to propagate and interact, leading to nonuniform spatial patterns.

\paragraph{(2) Damage dynamics:}
\begin{equation}
\frac{\partial D}{\partial t}
= -\beta D (1-h).
\label{eq:damage}
\end{equation}
The variable $D(x,t)$ represents the local density of damage. 
As the healing field increases, the damage field decreases proportionally, and the coefficient $\beta$ controls the decay rate. 
This antagonistic relationship ensures that regions with active recovery experience a gradual reduction in damage, generating smooth but spatially varying transitions between healed and wounded areas.

\paragraph{(3) Stress dynamics:}
\begin{equation}
\frac{\partial \sigma}{\partial t}
= k_1 h - k_2 D - k_3 \sigma.
\label{eq:stress}
\end{equation}
The variable $\sigma(x,t)$ denotes the local mechanical stress on the surface. 
The term $k_1 h$ models tension generated during healing, 
$-k_2 D$ reduces stress in damaged areas, 
and $-k_3\sigma$ represents relaxation over time. 
Although the stress field is not explicitly solved in the later diffusion experiments, it provides a means to modulate node positions during mesh generation. 
In this sense, stress acts as a spatial feedback term that perturbs the geometry in proportion to local healing and damage imbalances, giving rise to a physically meaningful irregular surface.

Although the stress field is neither solved together with the coupled healing–damage system 
nor used in the later diffusion experiments, 
it serves as a modulation factor for node displacement during mesh generation. 
Through this mechanism, stress provides spatial feedback that adjusts the geometry 
according to local healing and damage imbalance, 
allowing the mesh to acquire physically interpretable irregularity.

\subsubsection{Mesh Construction from PDE Dynamics}
\label{subsubsec:mesh_generation}

Our mesh\footnote{See GitHub: \url{https://github.com/abyssyli/2025Summer/blob/main/PINNs/Heading_and_Damage_PDE_Driven_Irregular_Mesh_visualization_and_code_analysis.ipynb}.} is generated through a simplified simulation of healing and damage processes on an ellipsoidal surface. 
The objective is to obtain an irregular geometry whose node distribution evolves under physical rules rather than artificial perturbations. 
The construction process involves three main stages: 
(1) sampling points on the ellipsoid, 
(2) evolving the surface through PDE dynamics, 
and (3) rebuilding the graph structure after the evolution stabilizes.

\paragraph{(1) Ellipsoid sampling}
The initial mesh consists of points sampled on the surface of an ellipsoid centered at $(0.5,0.5,0.5)$ with semi-axes $(a,b,c)=(0.6,0.4,0.3)$. 
The angular coordinates are drawn as
\begin{equation}
\theta = 2\pi u_1, \quad \phi = \arccos(2u_2 - 1),
\end{equation}
where $u_1,u_2 \sim U(0,1)$ are uniform random variables. 
The corresponding spatial coordinates are
\begin{equation}
\mathbf{x} =
\begin{bmatrix}
a \sin\phi \cos\theta,~
b \sin\phi \sin\theta,~
c \cos\phi
\end{bmatrix}^\top + \mathbf{c}.
\end{equation}
Sampling on a curved surface leads to a mild nonuniformity in point density, which reflects local curvature.

\paragraph{(2) PDE-based surface evolution}
Each sampled node carries two scalar variables, healing $h_i$ and damage $D_i$, 
which evolve according to the coupled system introduced in Section~3.1.2 (Eqs.~\ref{eq:healing}--\ref{eq:stress}). 
These equations are integrated numerically over discrete time steps using a graph Laplacian operator. 
The diffusion term controls spatial smoothing of $h$, 
and the reaction terms determine how local recovery fronts develop on the surface. 
The resulting fields $\{h_t, D_t, \sigma_t\}$ modify the spatial distribution of nodes and encode gradual geometric variation.

\paragraph{(3) Stress modulation and node displacement}
During each update step, the auxiliary stress field $\sigma_t$ provides a spatial correction that slightly adjusts node positions based on the local balance between healing and damage. 
This generates moderate geometric variation while maintaining overall surface continuity. 
The deformation does not alter mesh connectivity but results in a nonuniform node distribution consistent with the simulated physical fields.

\paragraph{(4) Rebuilding the graph structure from evolved nodes}
After the PDE evolution reaches a steady state, the node coordinates are fixed. 
Since the healing–damage process and stress modulation have slightly changed node positions, 
the spatial neighborhood relations must be redefined. 
This step rebuilds the graph structure by connecting each node to its $k$ nearest neighbors ($k=10$), 
forming an adjacency matrix $A$ that represents the final topology of the surface. 
From $A$, a discrete Laplacian operator is computed as
\begin{equation}
L = D^{-1}(A - D),
\end{equation}
where $D$ is the diagonal degree matrix. Here, $L = D^{-1}(A - D)$ follows the \textit{generator convention}, 
which ensures that each row of $L$ sums to zero and that all eigenvalues are non-positive. 
Under this convention, the discrete diffusion equation takes the form 
\begin{equation}
\frac{\partial u}{\partial t} = L u,
\end{equation}
so that $L$ acts as the discrete diffusion operator (the negative of the standard graph Laplacian).

This Laplacian serves as the diffusion stencil for both the Crank–Nicolson baseline and the CoreGNN model. 
Rebuilding the graph structure ensures that the final irregular mesh, after physical evolution, 
can be directly used for numerical diffusion and graph-based message passing.

\begin{figure}[htbp]
    \centering
    \begin{subfigure}[t]{0.49\linewidth}
        \centering
        \includegraphics[width=\linewidth]{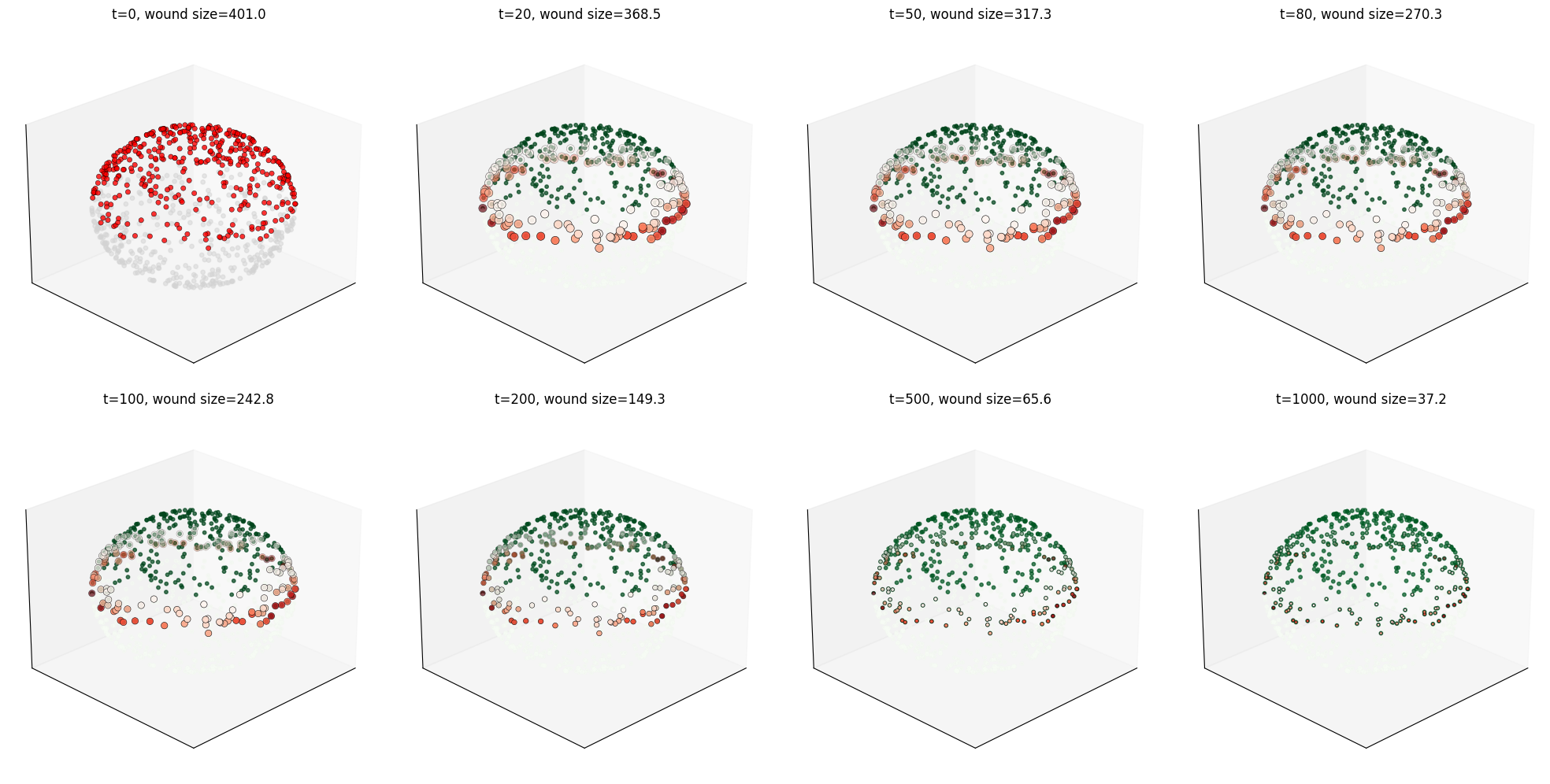}
        \caption{Evolution of the physically driven irregular mesh showing spatial healing–damage dynamics on the ellipsoidal surface.}
        \label{fig:healing_damage_mesh}
    \end{subfigure}
    \hfill
    \begin{subfigure}[t]{0.49\linewidth}
        \centering
        \includegraphics[width=\linewidth]{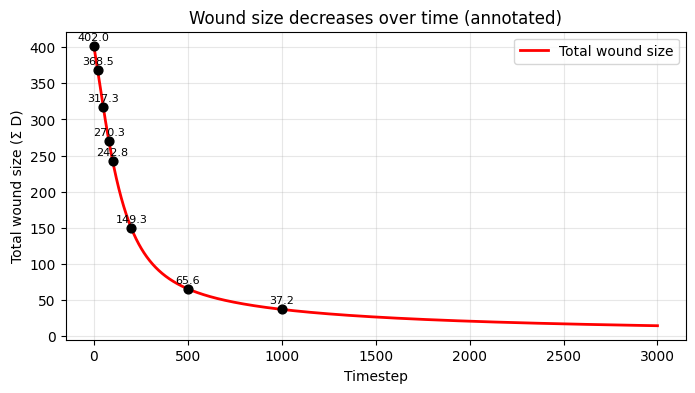}
        \caption{Temporal decay of total wound size $\sum D$, demonstrating the gradual healing process over simulation time.}
        \label{fig:wound_size}
    \end{subfigure}

    \caption{Physically driven mesh generation and wound healing dynamics. 
    (a)~Spatial evolution of the healing–damage process on the ellipsoidal mesh. 
    (b)~Decay of total wound size over time, reflecting the convergence toward a healed steady state.}
    \label{fig:mesh_evolution_combined}
\end{figure}

The resulting evolution of the healing–damage process and its geometric effects 
are illustrated in Figure~\ref{fig:mesh_evolution_combined}. 
The left panel shows how damage (red) gradually decreases as healing (green) propagates across the ellipsoidal surface, 
while the right panel presents the corresponding decay of total wound size over time. 
Together, these results confirm that the generated mesh captures physically consistent irregularity 
arising from PDE-driven evolution.

\subsubsection{Data Generation for Training}

Following the physically driven mesh construction described in Section~3.1.3,
the resulting node coordinates, connectivity, and physical variables $(h, D, \sigma)$
are transformed into structured graph samples for training.
Each mesh snapshot is stored as a \texttt{PyTorch Geometric} \texttt{Data} object
that includes the node positions $\mathbf{x}_i \in \mathbb{R}^3$, 
edge indices $\mathcal{E}$ defined by $k$-nearest neighbors, 
and boundary nodes $\mathcal{B}$ determined from the ellipsoidal surface equation. 
The final healing field $h$ is used as the initial condition $u_0$ 
for the diffusion tasks introduced in Section~3.2.

Edge weights are defined inversely proportional to the Euclidean distance
between connected nodes,
\begin{equation}
w_{ij} = \frac{1}{\lVert \mathbf{x}_i - \mathbf{x}_j \rVert_2 + \epsilon},
\label{eq:edge weight}
\end{equation}
ensuring that the diffusion operator remains consistent with the underlying geometry.
Boundary nodes are detected using the analytical ellipsoid constraint,
\begin{equation}
\frac{(x - c_x)^2}{a^2} + \frac{(y - c_y)^2}{b^2} + \frac{(z - c_z)^2}{c^2} = 1,
\end{equation}
with a small tolerance $\delta$ to assign Dirichlet or mixed-type boundary conditions for solving the PDEs.

The resulting dataset forms a collection of physically meaningful graph samples
defined on irregular surfaces, which are used as input domains for both
the Crank–Nicolson baseline and CoreGNN model training in the following sections.

\subsection{Real-World Irregular Meshes}

In addition to synthetic perturbations and the physically driven meshes
introduced previously, we also incorporate irregular meshes obtained directly
from real-world objects. Natural surfaces exhibit geometric variability that
cannot be reproduced by random noise or by simplified physical evolution
models. Such variability arises from material growth, anisotropy, scanning
artifacts, and structural heterogeneity, providing a complementary source of
irregularity for evaluating PDE solvers on complex domains.

\subsubsection{Motivation}

Many surfaces in nature, including leaves, shells, and biological tissues,
present nonuniform sampling density, curvature concentration, fine geometric
details, and open boundaries. These characteristics appear naturally during
growth or formation and cannot be captured by simple perturbations applied to
uniform grids. Similarly, high-resolution scanned models such as the Stanford
Bunny contain anisotropic neighborhoods and detailed local structures. These
real-world meshes serve as a critical benchmark to determine whether
graph-based PDE solvers remain stable and accurate on domains whose
irregularity is not artificially generated.

\subsubsection{Mesh Acquisition and Processing}

To construct real-world irregular meshes for PDE learning, we use three
representative surfaces: a scanned leaf, a scanned coconut fragment, and the
Stanford Bunny model. Although their sources differ, each mesh follows a
unified preprocessing pipeline to ensure consistent representation.

\begin{itemize}
    \item \textbf{Geometry preparation.}
    Each mesh is recentered by subtracting the vertex-wise mean. For scanned
    objects with a preferred orientation, a fixed rotation is applied so that
    the geometry aligns with a common reference frame.

    \item \textbf{Sampling of surface points.}
    When the original mesh contains more vertices than required, a uniform set
    of $N$ surface points is selected without replacement. No artificial
    resampling, remeshing, or smoothing is introduced, preserving the inherent
    irregularity of the object.

    \item \textbf{Graph construction.}
    Node coordinates in $\mathbb{R}^3$ are used to build a $k$-nearest-neighbor
    graph with a fixed neighborhood size. Edge weights follow a geometry-aware
    rule:

    \begin{equation}
        w_{ij} = \frac{1}{\lVert x_i - x_j \rVert_2 + 10^{-6}}.
    \end{equation}

    \item \textbf{Boundary detection.}
    Boundary vertices are identified by locating edges that belong to only one
    triangular face in the original mesh. These nodes are assigned as boundary
    points for the diffusion tasks.

    \item \textbf{Initial condition.}
    A smooth Gaussian field centered at the geometric mean is used:

    \begin{equation}
        u_0(x_i) = 
        \exp\!\left(
            -20 \,
            \frac{\lVert x_i - \bar{x} \rVert_2^2}
                 {\max_j \lVert x_j - \bar{x} \rVert_2^2}
        \right).
    \end{equation}

    \item \textbf{Diffusivity.}
    A homogeneous scalar diffusivity is assigned to all nodes:

    \begin{equation}
        D(x_i) = 0.05.
    \end{equation}

\end{itemize}

\subsubsection{Conversion to Graph-Based PDE Data}

After preprocessing, each real-world mesh is converted into a standardized
graph sample suitable for PDE simulation and GNN-based diffusion modeling.
The three-dimensional coordinates serve as node features, while the $k$-NN
edges define the spatial coupling required by the discrete diffusion operator.
Boundary nodes receive Dirichlet conditions, interior nodes evolve freely, and
the weighted edges determine the effective diffusion stencil. Each mesh
instance is stored as a graph containing:

\begin{itemize}
    \item node coordinates $x_i \in \mathbb{R}^3$,
    \item edge index pairs describing the $k$-nearest-neighbor connectivity,
    \item geometric edge weights $w_{ij}$,
    \item a boundary mask identifying fixed-value nodes,
    \item the initial condition $u_0(x_i)$,
    \item the diffusivity field $D(x_i)$.
\end{itemize}

This representation allows both numerical solvers and graph-based models to
operate on the same discretized domain. The resulting dataset therefore
provides a consistent, geometry-aware collection of irregular surfaces for
evaluating CoreGNN and baseline methods under realistic geometric complexity.

\subsubsection{Representative Surfaces}

\textbf{Leaf surface.}
This mesh originates from a physical scan of a plant leaf. Its geometry
contains natural anisotropy, curvature ridges, and open boundary curves,
reflecting biological structure rather than synthetic perturbation.

\textbf{Coconut fragment.}
The coconut shell exhibits a highly irregular, curved surface with strong
geometric variation across its boundary region. The surface roughness and
nonuniform sampling provide a challenging domain for diffusion dynamics.

\textbf{Stanford Bunny.}
This model contains complex folds and narrow regions and serves as a
high-resolution real-world mesh. Although widely used, its irregular sampling
makes it a meaningful benchmark for stability and consistency of PDE learning
methods.

Real-world meshes therefore complement synthetic and physically driven
geometries, allowing us to evaluate graph-based PDE solvers under naturally
occurring irregularity.

\begin{figure}[htbp]
    \centering

    \begin{subfigure}[t]{0.32\linewidth}
        \centering
        \includegraphics[width=\linewidth]{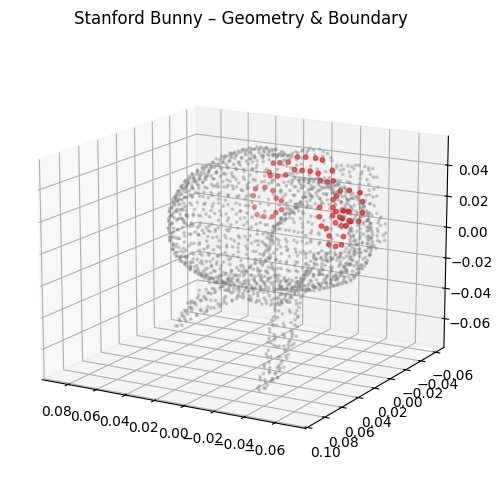}
        \caption{Stanford Bunny: geometry and detected boundary nodes.}
        \label{fig:bunny_mesh}
    \end{subfigure}
    \hfill
    \begin{subfigure}[t]{0.32\linewidth}
        \centering
        \includegraphics[width=\linewidth]{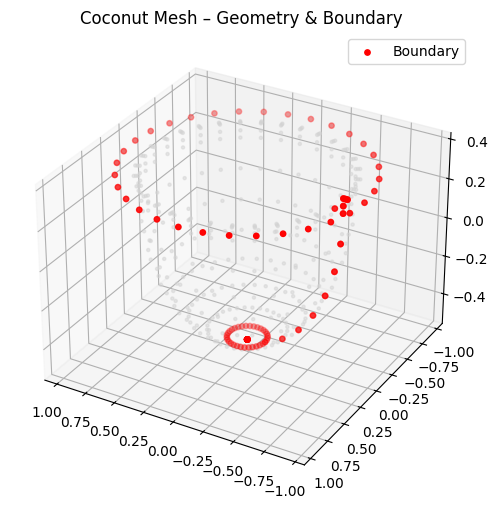}
        \caption{Coconut mesh: irregular curvature and boundary loops.}
        \label{fig:coconut_mesh}
    \end{subfigure}
    \hfill
    \begin{subfigure}[t]{0.32\linewidth}
        \centering
        \includegraphics[width=\linewidth]{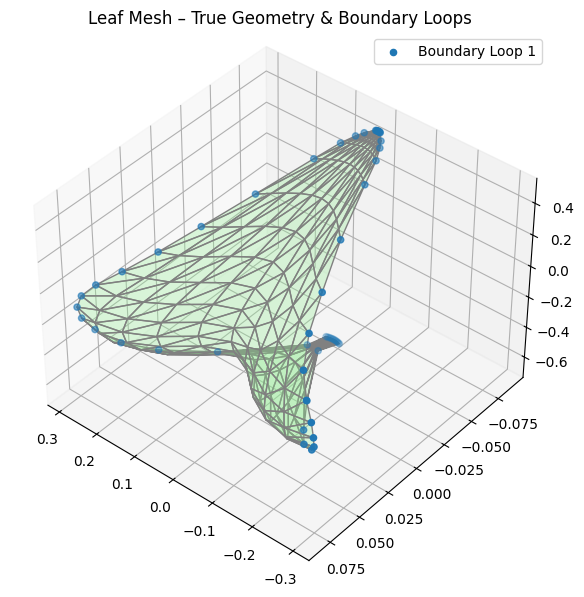}
        \caption{Leaf surface: true geometry with extracted boundary loops.}
        \label{fig:leaf_mesh}
    \end{subfigure}

    \caption{Real-world irregular meshes used in our dataset.  
    Each surface exhibits naturally occurring geometric irregularity, including 
    nonuniform sampling, complex curvature, and open boundaries, providing 
    challenging domains for PDE-based diffusion learning.}
    \label{fig:real_world_meshes}
\end{figure}

\section{Model Formulation}

\subsection{Problem Definition}

We consider the diffusion of a scalar field $u(\mathbf{x}, t)$ defined on a set of spatial points sampled from a three-dimensional ellipsoidal domain,
\begin{equation}
\Omega = \{(x,y,z) \mid (x-0.5)^2/a_0^2 + (y-0.5)^2/b_0^2 + (z-0.5)^2/c_0^2 \le 1\}.
\end{equation}
Each node $i$ in the irregular mesh corresponds to a spatial location $\mathbf{x}_i \in \Omega$ and is connected to its neighbors $\mathcal{N}(i)$ through weighted edges.  
Let $D_i$ denote the local diffusion coefficient at node $i$.  
The discrete diffusion dynamics on the graph are expressed as
\begin{equation}
\frac{\partial u(\mathbf{x}_i, t)}{\partial t}
= D_i \sum_{j \in \mathcal{N}(i)} w_{ij}\,[u(\mathbf{x}_j, t) - u(\mathbf{x}_i, t)],
\label{eq:discrete_diffusion}
\end{equation}
where $w_{ij}$ denotes the edge weight between nodes $i$ and $j$, following \ref{eq:edge weight}, and $u(\mathbf{x}_i, t)$ is equivalent to the node variable $u_i(t)$ representing the scalar field value at node $i$.  
This discrete form captures how local differences in $u$ drive diffusive fluxes across neighboring nodes of the irregular mesh.

In the continuous formulation, Eq.~\eqref{eq:discrete_diffusion} corresponds to the standard heterogeneous diffusion equation,
\begin{equation}
\frac{\partial u(\mathbf{x}, t)}{\partial t}
= D(\mathbf{x})\,\nabla^2 u(\mathbf{x}, t),
\label{eq:continuous_diffusion}
\end{equation}
where $D(\mathbf{x})$ is a spatially varying diffusivity field.  
Dirichlet boundary conditions are imposed on the ellipsoidal surface $\partial \Omega$ as
\begin{equation}
u(\mathbf{x}, t) = 0, \qquad \mathbf{x} \in \partial\Omega,
\end{equation}
and the initial condition is specified by a precomputed field $u_0(\mathbf{x})$:
\begin{equation}
u(\mathbf{x}, 0) = u_0(\mathbf{x}).
\end{equation}

This formulation defines the diffusion process to be learned by the proposed CoreGNN model, which approximates the discrete diffusion operator in Eq.~\eqref{eq:discrete_diffusion} through message passing on irregular meshes and models the temporal evolution of $u(\mathbf{x}, t)$ as a physically consistent graph-based solver for Eq.~\eqref{eq:continuous_diffusion}.

\subsection{overview of the OCGNN-PINN Model}
The proposed \textbf{Operator-Consistent Graph Neural Network (OCGNN-PINN)} is developed to approximate the diffusion operator on irregular meshes by embedding geometric relations directly into the message-passing process. 
Unlike conventional PINNs or supervised neural solvers, our model does not rely on labeled data or separate PDE training stages. 
Instead, it integrates the residual formulation of physics-informed neural networks into the graph neural network model structure, where the governing equation, boundary condition, and initial condition are all enforced through three residual-based losses. 
In addition, the design is conceptually inspired by the P-tensor representation, which introduces a node–edge duality reflecting the physical coupling between field values and fluxes. 
Given node features corresponding to the physical state $u(\mathbf{x}_i, t)$, the OCGNN-PINN learns the temporal evolution of the field through node–edge–edge–node propagation, allowing it to serve as a graph-based surrogate solver for the diffusion equation.

\subsection{Layer Design}

The model is organized around a Node–Edge–Edge–Node propagation structure, 
which is designed to describe how local field information is transferred and reorganized across the graph.  
Each propagation cycle begins from node representations that store scalar field values, 
followed by a node-to-edge transformation where geometric relations such as edge length, direction, 
and angular orientation are embedded to represent the spatial organization of the irregular mesh.  
Adjacent edges sharing a node then exchange information through an edge-to-edge refinement step, 
allowing the model to represent local spatial continuity without forming dense connectivity.  
The updated edge states are finally aggregated back to the node level, completing one full pass of information flow.  

The \textit{CoreGNNLayer} serves as the basic computational unit for this process.  
Multiple layers are stacked to form the \textit{CoreGNN} backbone, which progressively integrates 
geometric and topological information over multiple propagation steps.  
Building upon this foundation, the \textit{CoreGNN\_PINN\_Improved}\footnote{Note that CoreGNN, CoreGNN\_improved, are the names of the models that we used in the experiment. They are all for the OCGNN-PINN approach/model.} model further introduces 
a Rational Fourier time encoder, an explicit edge-dynamics head, and a learnable anchor correction module 
that improves the stability of Laplacian-based updates.

\subsection{Time-Dependent Dynamics}
\label{subsec:time_dynamics}

The diffusion process evolves over time according to the heterogeneous diffusion
equation introduced in Section~4.1.  
To approximate this evolution on irregular meshes, the proposed model adopts an
explicit discrete-time formulation.  
Instead of embedding time as a continuous input or learning a direct mapping
from $(x,t)$ to $u(x,t)$, the model predicts the instantaneous temporal derivative
of the field at each step and advances the solution using a neural time-marching
scheme.

\subsubsection{Discrete update of the diffusion field}

Let $f_t \in \mathbb{R}^N$ denote the vector of node values at time $t$, representing
the solution $u(x_i,t)$ evaluated at mesh nodes.  
Given a step size $\Delta t$, the temporal evolution is computed by an explicit
Euler update,
\begin{equation}
f_{t+\Delta t} = f_t + \Delta t\, \dot{f}(t),
\end{equation}
where $\dot{f}(t)$ is the instantaneous temporal derivative predicted by the
neural network.  
Because the network receives the current state $f_t$ at every step, it effectively
learns a state-dependent dynamics function
\begin{equation}
\dot{f}(t) = \Phi_{\theta}\!\left(f_t,\; \mathcal{G}\right),
\end{equation}
where $\mathcal{G}$ denotes the static graph structure of the irregular mesh and
$\Phi_{\theta}$ is implemented by the CoreGNN backbone described in
Section~4.3.  
This formulation mirrors standard explicit schemes used in numerical PDE solvers
and provides a stable mechanism for modeling temporal change.

\subsubsection{Sequential Time-Stepping}

During both training and inference, the model evolves the solution across a
sequence of discrete time points
\begin{equation}
t_0 < t_1 < \cdots < t_{n_t},
\end{equation}
starting from the prescribed initial condition $f_0 = u_0$.  
Sequential application of the discrete update generates the neural rollout
\begin{equation}
f_0 \rightarrow f_1 \rightarrow f_2 \rightarrow \cdots \rightarrow f_{n_t},
\end{equation}
which approximates the temporal trajectory of the diffusion field.  
This discrete-time dynamics framework enables the model to propagate information
through both time and space without requiring continuous-time embeddings,
auxiliary temporal encodings, or implicit integration schemes.

\subsection{Physics-Informed Training}
Following the physics-informed learning paradigm, the model is trained using
residual objectives derived from the diffusion equation together with boundary
and initial conditions. These components form the standard loss terms in our
framework. In addition, we incorporate a structure-preserving penalty motivated
by prior work such as Kütük and Yücel~\cite{kutuk2024allen}, who introduced an
energy-dissipation term to improve stability in Allen--Cahn PINNs. While their
approach enforces a physical energy law, our formulation instead imposes an
operator-level consistency constraint that links the discrete gradient and
divergence operators. This provides a complementary type of regularization that
is specifically suited to irregular meshes and helps stabilize the learned
dynamics by maintaining the algebraic structure of the underlying operators.

\subsection{Standard PINN Loss}
\paragraph{(i) PDE residual loss}
At each time $t$, the network predicts the instantaneous node dynamics $\dot f$.
The discrete diffusion residual is defined as
\begin{equation}
R_{\text{PDE}} = \dot f - D(Lf),
\label{eq:pde_residual}
\end{equation}
where $L$ denotes the normalized graph Laplacian and $D$ the per-node diffusivity. 
Minimizing $\|R_{\text{PDE}}\|_2$ drives the learned dynamics to satisfy the diffusion relation on the irregular mesh.
A dynamic scaling factor is further applied in implementation to balance the magnitudes of the temporal and spatial terms.

\paragraph{(ii) Boundary-condition loss}
For boundary nodes identified by mask $\mathcal{B}$,
a penalty term constrains the solution to satisfy Dirichlet boundary conditions,
where the field variable $u$ is fixed to a prescribed value on the boundary.
In this study, we set $u_i(t)=0$ for $i\in\mathcal{B}$,
representing a fixed zero-value boundary where the field vanishes at the domain edges.
This enforces the correct boundary behavior throughout training.

\paragraph{(iii) Initial-condition loss}
A regularization term ensures that the predicted field at $t=0$ matches the prescribed initial state $u_0$,
providing a stable reference for the temporal rollout.

The three losses are combined into a single training objective with adaptive weights
$(\lambda_{\text{PDE}},\lambda_{\text{BC}},\lambda_{\text{IC}})$
that evolve dynamically during optimization.
Following an inverse-logarithmic scheduling rule, larger residuals are automatically down-weighted when they dominate,
while smaller residuals gain relatively higher influence.
This adaptive balancing prevents the optimization from collapsing toward a single dominant constraint
and stabilizes convergence among physics, boundary, and initialization objectives.

\subsubsection{Incidence, Dynamics, Residuals, Conservation, and Permutation Equivariance}

\paragraph{Fields and definitions }
Let $f_t: \mathbb{V} \to\mathbb R$ be a node field and $g_t:\mathbb E\to\mathbb R$ an edge field.
Consider an edge-by-node incidence matrix $B\in\mathbb{R}^{E\times N}$.
Each undirected edge is assigned an arbitrary orientation $\mathrm{src}(e)\to\mathrm{dst}(e)$.
$\\$$\\$
With the incidence 
\begin{equation}
B[e,\mathrm{src}(e)] \;=\; +1,\qquad
B[e,\mathrm{dst}(e)] \;=\; -1,
\label{eq:inc}
\end{equation}
the edge ``gradient'' and node ``divergence'' read
\begin{equation}
(B f_t)[e] \;=\; f_t(\mathrm{src}(e)) - f_t(\mathrm{dst}(e)),
\qquad
(B^\top g_t)[v] \;=\; \sum_{e:\,v=\mathrm{src}(e)} g_t(e)\;-\!\!\!\!\!\!\!\!\!\sum_{e:\,v=\mathrm{dst}(e)} g_t(e),
\label{eq:local-ours}
\end{equation}
i.e., $(B^\top g_t)[v]=$ outflow $-$ inflow at node $v$, and we define the edge-to-node
divergence by
\begin{equation}
\bigl(\nabla g_t\bigr)[v]
\;:=\; \sum_{u\sim v} g_t((u,v)) \;-\; \sum_{u\sim v} g_t((v,u))
\;=\; \text{inflow} \;-\; \text{outflow}
\;=\; -\,(B^\top g_t)[v].
\label{eq:hand-sigma}
\end{equation}
$\\$$\\$
Here we use $\nabla:\mathbb{R}^{E}\to\mathbb{R}^{N}$ to denote the edge-to-node divergence
with the sign convention “inflow $-$ outflow”.
$\\$$\\$
Hence the two notations are exactly equivalent:
\begin{equation}
\partial_t f_t \;=\; B^\top g_t \;=\; -\,\nabla g_t .
\label{eq:dtf-bridge}
\end{equation}
$\\$$\\$
We will use the matrix form $\partial_t f_t=B^\top g_t$ in the sequel; all results
(energy conservation, permutation equivariance, etc.) are independent of this notational choice.

\paragraph{First-order dynamics}
We consider the skew-coupled first-order system
\begin{equation}
\dot f \;=\; B^\top g,\qquad
\dot g \;=\; -\,B f,
\label{eq:dyn}
\end{equation}
where $f(t)\in\mathbb{R}^N$ is a node field and $g(t)\in\mathbb{R}^E$ is an edge-aligned flux.

\paragraph{Residuals (left minus right)}
We define residuals in the standard \emph{left minus right} form:
\begin{equation}
R_f \;=\; \dot f \;-\; B^\top g,\qquad
R_g \;=\; \dot g \;+\; B f.
\label{eq:residuals}
\end{equation}
The ``$+$'' in $R_g$ is mandatory because the physical equation for $\dot g$ in \eqref{eq:dyn} carries a minus sign:
\begin{equation}
R_g \;=\; \dot g \;-\;(-B f) \;=\; \dot g \;+\; B f.
\label{eq:why-plus}
\end{equation}
Using a minus here would correspond to $\dot g=+Bf$ and would break energy conservation (see \eqref{eq:energy-deriv}).

\paragraph{Second-order forms}
Differentiating $\dot f = B^\top g$ and substituting $\dot g = -B f$ gives
\begin{equation}
\ddot f \;=\; B^\top \dot g \;=\; B^\top(-B f) \;=\; -\,B^\top B\, f .
\label{eq:second-f}
\end{equation}
Similarly, differentiating $\dot g = -B f$ and substituting $\dot f = B^\top g$ yields
\begin{equation}
\ddot g \;=\; -\,B \dot f \;=\; -\,B(B^\top g) \;=\; -\,B B^\top g .
\label{eq:second-g}
\end{equation}
We therefore identify the (symmetric positive semidefinite) node/edge Laplacians
\begin{equation}
L_V \;:=\; B^\top B \;\in\; \mathbb{R}^{N\times N},\qquad
L_E \;:=\; B B^\top \;\in\; \mathbb{R}^{E\times E}.
\label{eq:laplacians}
\end{equation}
Positivity follows from the quadratic forms
\begin{equation}
x^\top L_V x \;=\; x^\top B^\top B x \;=\; \|B x\|_2^2 \;\ge\; 0,\qquad
y^\top L_E y \;=\; \|B^\top y\|_2^2 \;\ge\; 0.
\label{eq:psd}
\end{equation}

\paragraph{Energy (Hamiltonian) conservation}
Define the quadratic Hamiltonian
\begin{equation}
\mathcal{H}(t) \;:=\; \tfrac12 \|f(t)\|_2^2 \;+\; \tfrac12 \|g(t)\|_2^2
\;=\; \tfrac12 f^\top f \;+\; \tfrac12 g^\top g .
\label{eq:energy}
\end{equation}
Differentiating and substituting \eqref{eq:dyn} yields
\begin{equation}
\dot{\mathcal{H}} \;=\; f^\top \dot f \;+\; g^\top \dot g
\;=\; f^\top B^\top g \;-\; g^\top B f
\;=\; (B f)^\top g \;-\; g^\top (B f) \;=\; 0,
\label{eq:energy-deriv}
\end{equation}
so $\mathcal{H}(t)$ is conserved along trajectories of \eqref{eq:dyn}. 
$\\$$\\$
If one incorrectly enforced $\dot g=+B f$, then
\begin{equation}
\dot{\mathcal{H}} \;=\; f^\top B^\top g \;+\; g^\top B f \;=\; 2\,g^\top B f \;\neq\; 0\ \text{ in general,}
\label{eq:wrong-sign}
\end{equation}
thus breaking conservation.

\paragraph{Permutation equivariance}
Let $P_V\in\mathbb{R}^{N\times N}$ be any node permutation matrix and
$Q_E\in\mathbb{R}^{E\times E}$ any \emph{signed} edge permutation (a permutation possibly
composed with edge-orientation flips). Both are orthogonal:
\begin{equation}
P_V^\top P_V = I_N,\qquad Q_E^\top Q_E = I_E.
\label{eq:perm-orth}
\end{equation}
Under relabeling we transform the variables as
\begin{equation}
f' = P_V f,\qquad g' = Q_E g,
\label{eq:perm-vars}
\end{equation}
and the incidence by the natural rule
\begin{equation}
B' = Q_E\, B\, P_V^\top \qquad \bigl(B\in\mathbb{R}^{E\times N}\bigr).
\label{eq:perm-B}
\end{equation}
Then the transformed variables satisfy the same-form dynamics with $B'$:
\begin{equation}
\dot f' = P_V \dot f = P_V B^\top g
= P_V B^\top Q_E^\top (Q_E g)
= (Q_E B P_V^\top)^\top g' = (B')^\top g',
\label{eq:perm-eq-f}
\end{equation}
\begin{equation}
\dot g' = Q_E \dot g = Q_E(-B f)
= -\,Q_E B P_V^\top (P_V f) = -\,B' f'.
\label{eq:perm-eq-g}
\end{equation}
In particular, if the initial data are transformed consistently,
\begin{equation}
f'(0) = P_V f(0),\qquad g'(0) = Q_E g(0),
\label{eq:perm-ic}
\end{equation}
then by uniqueness of solutions to linear ODEs we obtain the flow-level equivariance:
\begin{equation}
f'(t) = P_V f(t),\qquad g'(t) = Q_E g(t)\qquad \forall\,t\ge 0.
\label{eq:flow-equiv}
\end{equation}
$\\$
Moreover, by \eqref{eq:perm-orth} we have $\|f'\|_2=\|f\|_2$ and $\|g'\|_2=\|g\|_2$, hence
\begin{equation}
\mathcal{E}'(t) \;=\; \tfrac12 \|f'\|_2^2 + \tfrac12 \|g'\|_2^2
\;=\; \tfrac12 \|f\|_2^2 + \tfrac12 \|g\|_2^2 \;=\; \mathcal{E}(t).
\label{eq:energy-invariance}
\end{equation}

\paragraph{Remark on edge orientation}
Flipping all edge orientations corresponds to replacing $B$ and $g$ by
$S_E B$ and $S_E g$, where $S_E\in\mathbb{R}^{E\times E}$ is diagonal with entries $\pm 1$.
Since $S_E^\top S_E=I_E$, both \eqref{eq:dyn} and \eqref{eq:energy-deriv} are unchanged, and so are the proofs above.
$\\$$\\$
Equivalently, the orientation flip can be summarized in compact form. 
Let $S_E=\mathrm{diag}(\pm1)$; then

\begin{equation}
B \mapsto S_E B,\quad g \mapsto S_E g,\qquad
\dot f = (S_E B)^\top (S_E g) = B^\top g,\quad
\dot g = - (S_E B) f = - B f .
\label{eq:orientation-inv}
\end{equation}
$\\$
Because $S_E^\top S_E=I_E$, \eqref{eq:orientation-inv} leaves the dynamics $\dot f=B^\top g$, $\dot g=-Bf$,
the Laplacians $\Delta_V=B^\top B$, $\Delta_E=BB^\top$, the Hamiltonian $\mathcal H$,
and the residual loss invariant; hence none of our results depends on the edge directions.
$\\$$\\$
Therefore, the subsequent results (energy conservation and permutation equivariance) do not depend on edge directions.

\paragraph{the P–tensor \cite{Hands2024Ptensors} consistency loss}

Having established the skew-coupled dynamics
$\dot f=B^\top g,\ \dot g=-Bf$,
energy conservation, and permutation equivariance, we encode these
\emph{structural} properties directly into training via a consistency loss.
Define the residuals (left minus right)
\begin{equation}
R_f \;:=\; \dot f - B^\top g,\qquad
R_g \;:=\; \dot g + B f.
\label{eq:ptensor-residuals}
\end{equation}
Our fourth loss penalizes their squared norms (time-averaged over sampled steps):
\begin{equation}
\mathcal{L}_{\mathrm{PT}} \;:=\;
\frac{1}{|\mathcal{T}|}\sum_{t\in\mathcal{T}}
\Big(\,\|R_f(t)\|_2^2 \;+\; \|R_g(t)\|_2^2\,\Big).
\label{eq:ptensor-loss}
\end{equation}
$\\$
We construct this P-tensor consistency loss for three reasons.
$\\$$\\$
Firstly, from\eqref{eq:ptensor-residuals} we have the  identity
\begin{equation}
\frac{d}{dt}\,\mathcal{H}(t)
= f^\top \dot f + g^\top \dot g
= f^\top R_f + g^\top R_g,
\label{eq:energy-drift}
\end{equation}
so $\mathcal{L}_{\mathrm{PT}}$ explicitly bounds energy drift:
\begin{equation}
\big|\dot{\mathcal{H}}(t)\big|
\;\le\; \|f(t)\|_2\,\|R_f(t)\|_2 + \|g(t)\|_2\,\|R_g(t)\|_2.
\label{eq:energy-bound}
\end{equation}
Thus driving $\mathcal{L}_{\mathrm{PT}}$ small yields near-conservation of $\mathcal{H}$ in training.
$\\$$\\$
Secondly, under node/edge (signed) permutations $(P_V,Q_E)$ we showed
$R_f' = P_V R_f$ and $R_g'=Q_E R_g$, hence
\begin{equation}
\|R_f'\|_2^2+\|R_g'\|_2^2
= \|R_f\|_2^2+\|R_g\|_2^2,
\label{eq:ptensor-invariant}
\end{equation}
so $\mathcal{L}_{\mathrm{PT}}$ is permutation-invariant and does not depend on the
arbitrary choice of edge orientations.
$\\$$\\$
Finally, we want to provide a complement\footnote{To balance magnitudes we optionally rescale each term by empirical means,
e.g.\ use $\widehat R_f := R_f/\alpha_f,\ \widehat R_g := R_g/\alpha_g$ with
$\alpha_f,\alpha_g$ taken as running estimates of $\|\dot f\|_2$ and $\|Bf\|_2$,
and replace $R_f,R_g$ by $\widehat R_f,\widehat R_g$ in \eqref{eq:ptensor-loss}.
This prevents a single head from dominating the gradients.} to the first three losses. Our total objective is
\begin{equation}
\mathcal{L}
= \lambda_{\mathrm{PDE}}\mathcal{L}_{\mathrm{PDE}}
+ \lambda_{\mathrm{BC}}\mathcal{L}_{\mathrm{BC}}
+ \lambda_{\mathrm{IC}}\mathcal{L}_{\mathrm{IC}}
+ \lambda_{\mathrm{PT}}\mathcal{L}_{\mathrm{PT}},
\label{eq:total-loss}
\end{equation}
where $\mathcal{L}_{\mathrm{PDE}}$ enforces the target physics on $f$,
$\mathcal{L}_{\mathrm{BC}}/\mathcal{L}_{\mathrm{IC}}$ anchor boundary and initial states, respectively, and the new $\mathcal{L}_{\mathrm{PT}}$ ties the node–edge heads together through the correct
skew-symmetric(energy-conserving) coupling, thus improving stability and generalization under graph relabeling.

\section{Experiments and Results}
\subsection{Experimental Setup}
\subsubsection{Dataset}
\label{subsubsec:dataset}

The experiments are conducted on the irregular meshes introduced in 
Section~\ref{sec:irregular_mesh}, which include both the physically driven 
ellipsoidal surfaces generated through healing--damage dynamics and the 
three real-world scans.  
Each surface is converted into a \texttt{PyTorch Geometric} graph containing 
three-dimensional node coordinates, a $k$-nearest-neighbor connectivity 
structure, geometry-aware edge weights, and boundary nodes detected from 
the underlying surface geometry.  
This conversion yields a consistent graph representation that preserves the 
original irregularity of each domain while enabling direct use in graph-based 
PDE solvers.

For all meshes, the initial condition and diffusivity field follow the definitions 
established in Section~\ref{sec:irregular_mesh}.  
The resulting collection of graph samples forms a unified dataset in which 
each irregular surface is represented by identical structural components, 
allowing both numerical baselines and learning-based models to operate 
under comparable geometric discretizations.

\subsubsection{Time Encoding and Rollout}
\label{subsubsec:time_rollout}

To represent the temporal evolution of the diffusion field, each training step provides the network with
the current time variable~$t$. Temporal information is injected through a multi-frequency Rational
Fourier encoding, which maps~$t$ into a compact vector of bounded sinusoidal features,
\begin{equation}
\gamma(t) = 
\Big[
\sin\!\big(\tfrac{\omega t}{1 + \omega t}\big),~
\cos\!\big(\tfrac{\omega t}{1 + \omega t}\big)
\Big]_{\omega \in \text{multiple frequency scales}}.
\end{equation}
This encoding provides smoothly varying temporal components across several frequency bands and avoids
the divergence that appears in standard sinusoidal encodings for large~$t$.

During both training and inference, the model outputs the predicted node state~$u_t$ together with its
time derivative~$\dot{f}_t$. Time evolution is generated through an explicit multi-step rollout. Given the
current state~$f_t$, the next-step value is obtained via a first-order update,
\begin{equation}
f_{t+\Delta t} = f_t + \Delta t \, \dot{f}_t,
\end{equation}
after which the new state~$f_{t+\Delta t}$ is fed back into the network for the next step. Repeating this
procedure produces the temporal trajectory $\{f_t\}_{t=0}^{T}$ starting from the initial condition~$u_0$.

This rollout mechanism is used exclusively at the experimental level to produce multi-step predictions
and does not modify the model architecture or the PDE formulation introduced in Section~4.

\subsubsection{Baselines}
\label{subsubsec:baseline}
To provide a physically grounded reference for evaluation, two categories of baselines are considered: 
(1)~numerical solvers that serve as ground truth for continuous diffusion trajectories, 
and (2)~neural models that approximate the same dynamics on irregular meshes.

We select Crank-Nicolson as our primary numerical reference because it operates directly on the discrete graph Laplacian constructed from our irregular meshes, avoiding the additional complexity and potential projection errors that would arise from applying finite element solutions computed on auxiliary meshes to our two-dimensional surfaces embedded in three-dimensional space.

\paragraph{Numerical reference (CN)}\footnote{See GitHub: \url{https://github.com/abyssyli/2025Summer/blob/main/PINNs/CN_Baselines_double_check_and_explanation.ipynb}}.

The Crank–Nicolson (CN) scheme~\cite{Guo2017CNStability} is used as the numerical baseline. 
CN is a standard implicit discretization method for diffusion-type partial differential equations, 
which integrates the temporal derivative using a midpoint (semi-implicit) rule to obtain a stable numerical solution. 
$\\$$\\$
It is solved on the same irregular mesh and diffusion coefficient field as the learning models, 
thereby providing a numerical reference under identical initial and boundary conditions.
$\\$
The discrete update can be written as
\begin{equation}
(I - \tfrac{\Delta t}{2}L)u_{t+\Delta t}
= (I + \tfrac{\Delta t}{2}L)u_t,
\label{eq:cn_update}
\end{equation}
where $L = D - A$ is the discrete graph Laplacian, 
defined with positive diagonal and negative off-diagonal entries following the standard sign convention in graph-based discretizations.
$\\$$\\$
Two CN variants are considered:
$\\$
(1)~\textbf{CN-irregular}, which employs a normalized graph Laplacian with edge weights $w_{ij}=1/d_{ij}$~\cite{Hein2007Laplacian}, and
(2)~\textbf{CN-PDE}\footnote{CN\text{-}PDE uses the physically scaled weights $1/d_{ij}^2$, which match the continuous diffusion operator. 
Because this scaling amplifies small geometric distortions, the resulting errors fluctuate more strongly across irregular meshes, 
so CN\text{-}PDE serves mainly as a supplementary diagnostic rather than a primary accuracy criterion.}
, which adopts strict $1/d_{ij}^2$~\cite{Guo2017CNStability} weights corresponding to the continuous diffusion operator.

\paragraph{Neural baselines}\footnote{See GitHub: \url{https://github.com/abyssyli/2025Summer/blob/main/PINNs/Baseline_comparisons_0920.ipynb} and \url{https://github.com/abyssyli/2025Summer/blob/main/PINNs/experiment10020203(baselines_comparison).ipynb}.}

In addition to the numerical reference, several neural models are trained under the same PDE, BC, and IC loss terms:

\begin{itemize}
    \item \textbf{MLP}: a coordinate-based baseline that directly maps spatial positions and initial values to predicted node states without using graph connectivity.
    \item \textbf{CNN}: a convolutional model trained on uniformly discretized 3D grids derived from the same spatial domain, 
    serving as a structured-grid reference for comparison.
    \item \textbf{GCN-PINN}: a spectral graph-convolution model equipped with Rational Fourier time encoding, allowing continuous-time inputs. 
    It models node-level diffusion dynamics through message passing on the irregular mesh but does not include explicit edge states.
\end{itemize}

These baselines collectively provide both numerical and neural references for evaluating the accuracy, stability, 
and physical consistency of our proposed OCGNN-PINN\footnote{In the experiment code on GitHub, our model was called CoreGNN, because the idea of operator consistency didn't come from the beginning} model.

\subsubsection{Evaluations}\label{sec:evaluations}

Unless otherwise specified, all errors are computed against the numerical reference (default: CN-irregular) at the final time step $T$, with temporal curves additionally reported to characterize long-term behavior. 
Let $u^{(\mathrm{pred})}(t)\in\mathbb{R}^N$ and $u^{(\mathrm{ref})}(t)\in\mathbb{R}^N$ denote the predicted and reference node states, respectively, $\mathcal{L}$ the discrete Laplacian operator, $D=\mathrm{diag}(D_1,\dots,D_N)$ the local diffusivity, and $\Delta t=T/n_t$ the time step size.

\paragraph{(1) Mean Absolute Error (MAE)}
The average absolute deviation at the final time:
\begin{equation}
\mathrm{MAE}(T)=\frac{1}{N}\sum_{i=1}^N \bigl|u^{(\mathrm{pred})}_i(T)-u^{(\mathrm{ref})}_i(T)\bigr|.
\end{equation}
This measure provides a scale-independent measure of overall deviation and is less sensitive to outliers.

\paragraph{(2) Mean Squared Error (MSE)}
The mean squared difference at the final time:
\begin{equation}
\mathrm{MSE}(T)=\frac{1}{N}\sum_{i=1}^N \bigl(u^{(\mathrm{pred})}_i(T)-u^{(\mathrm{ref})}_i(T)\bigr)^2.
\end{equation}
MSE emphasizes larger deviations and aligns with the quadratic loss used during training.

\paragraph{(3) $L_2$ Solution Error}
A normalized $L_2$ error is reported as
\begin{equation}
\|e(T)\|_{2,\mathrm{norm}}
=\frac{\bigl\|u^{(\mathrm{pred})}(T)-u^{(\mathrm{ref})}(T)\bigr\|_2}{\sqrt{N}},
\end{equation}
which allows scale-consistent comparison across meshes of different sizes. The temporal evolution $\|e(t)\|_{2,\mathrm{norm}}$ further reflects the stability of each method over time.

\paragraph{(4) PDE Residual}
The physics-based residual quantifies the deviation from the diffusion equation:
\begin{equation}
R(t)=\widehat{\partial_t u}(t)-D\,\mathcal{L}\,u^{(\mathrm{pred})}(t),
\end{equation}
where $\widehat{\partial_t u}(t)$ is approximated by a first-order forward difference.
The time-averaged residual magnitude is defined as
\begin{equation}
\|R\|_{\mathrm{time}}
=\frac{1}{\sqrt{n_t N}}\Biggl(\sum_{k=0}^{n_t-1}\bigl\|R(t_k)\bigr\|_2^2\Biggr)^{1/2},
\end{equation}
which is reported as the \emph{PDE residual}, representing the overall deviation from physical consistency across the temporal window.

\paragraph{(5) Loss Convergence}
Training and validation losses, including PDE, boundary, and initial condition components, are tracked over epochs on a logarithmic scale. 
These curves reveal the optimization stability and convergence behavior of each model, complementing the final error evaluations.

\subsection{Results on Synthetic and Physically Driven Meshes}
\label{subsec:synthetic_results}

We conducted a comprehensive series of experiments across three categories of irregular
geometries: the \textbf{physically driven ellipsoidal meshes}, the \textbf{Stanford Bunny} surface,
and two additional \textbf{real-world scanned meshes} (leaf and coconut).
These domains exhibit distinct curvature profiles, sampling nonuniformity, and boundary
structures, enabling a broad evaluation of the proposed model under diverse geometric
irregularity. Our study systematically compares the baseline \textbf{CoreGNN (v1.0)} with the enhanced
\textbf{CoreGNN\_Improved (v2.0)}.
The transition from v1.0 to v2.0 incorporates a sequence of targeted architectural and
training refinements, including \textbf{anchor correction}, explicit \textbf{boundary handling},
\textbf{diffusivity calibration}, \textbf{dynamic loss weighting}, \textbf{loss scaling},
\textbf{synchronized rollout updates}, \textbf{continuous time encoding}, and the new
\textbf{P-tensor Consistency Loss}.
To assess the contribution of each component, the experiments include both 
\textbf{horizontal comparisons} (across model architectures under identical conditions)
and \textbf{vertical comparisons} (examining the cumulative effect of incremental design
modifications on accuracy and stability).

\subsubsection{Comparison 1: Random Ellipsoid Point-Cloud Benchmark}

We first performed a horizontal comparison on a randomly sampled ellipsoidal point-cloud mesh,
contrasting the proposed CoreGNN-PINN with several representative baselines
(GCN, PINN, CNN, and MLP) trained under identical physical and numerical settings\footnote{See GitHub: \url{https://github.com/abyssyli/2025Summer/blob/main/PINNs/Baseline_comparisons_0920.ipynb}}.

\begin{table}[H]
\centering
\caption{Random ellipsoidal point-cloud benchmark: horizontal comparison across architectures.
All models are trained under identical physical and numerical settings.}
\label{tab:comp1}
\begin{tabular}{lcc}
\toprule
\textbf{Model} & \textbf{MAE (CN baseline)} & \textbf{MSE (CN baseline)} \\
\midrule
CoreGNN-PINN & $\bm{2.756\times10^{-2}}$ & $8.875\times10^{-3}$ \\
GCN-PINN     & $3.278\times10^{-2}$ & $\bm{6.018\times10^{-3}}$ \\
PINN (MLP + Physics) & $6.342\times10^{-2}$ & $1.517\times10^{-2}$ \\
CNN           & $7.593\times10^{-2}$ & $5.525\times10^{-2}$ \\
MLP (No Physics) & $1.400\times10^{-1}$ & $4.156\times10^{-2}$ \\
\bottomrule
\end{tabular}
\end{table}

\subsubsection{Comparison 2: Damage--Healing Mesh Benchmark}

While the previous comparison focused on the random ellipsoidal geometry,
this experiment evaluates the \textbf{CoreGNN\_Improved} model on a
\textbf{2D damage--healing mesh embedded in a 3D surface}.
The damage--healing mesh setting represents a static reconstruction task,
where the model learns the spatial distribution of the diffusive field
over an irregular 3D surface with heterogeneous coefficients.
This configuration serves as a more challenging and physically consistent
benchmark than the random ellipsoidal point-cloud case.

\begin{table}[H]
\centering
\caption{Damage--healing mesh benchmark: comparison of CoreGNN and GCN against CN baselines
under normalized and PDE-scaled evaluation metrics.}
\label{tab:comp2}
\begin{tabular}{lcccc}
\toprule
\textbf{Metric Type} & \textbf{Model} & \textbf{MAE} & \textbf{MSE} & \textbf{L$_2$ norm} \\
\midrule
CN\_norm (normalized) &
CoreGNN & $\bm{2.0127\times10^{-2}}$ & $\bm{2.0330\times10^{-3}}$ & $\bm{4.5089\times10^{-2}}$ \\
CN\_norm (normalized) &
GCN & $3.1958\times10^{-1}$ & $1.2294\times10^{-1}$ & $3.5063\times10^{-1}$ \\
CN\_PDE ($1/d^2$ weight) &
CoreGNN & $\bm{3.4762\times10^{-1}}$ & $2.4048\times10^{-1}$ & --- \\
CN\_PDE ($1/d^2$ weight) &
GCN & $3.7492\times10^{-1}$ & $\bm{1.6309\times10^{-1}}$ & --- \\
\midrule
\textbf{PDE Residual $L_2$} &
CoreGNN & \multicolumn{3}{c}{$\bm{5.9005\times10^{-2}}$} \\
\textbf{PDE Residual $L_2$} &
GCN & \multicolumn{3}{c}{$4.7730\times10^{-1}$} \\
\bottomrule
\end{tabular}
\end{table}

\noindent
The results highlight that the \textbf{CoreGNN\_Improved} model \footnote{See GitHub: \url{https://github.com/abyssyli/2025Summer/blob/main/PINNs/experiment100601(the_best_ever).ipynb}.}substantially outperforms the GCN baseline
across all evaluation criteria. In particular, its PDE residual norm is nearly an order of magnitude lower,
demonstrating superior enforcement of physical consistency under heterogeneous diffusivity and
nonlinear boundary constraints. The gains are consistent across both normalized and PDE-scaled
evaluations, indicating improved numerical stability in long-time rollouts.

\subsubsection{Loss and Physical Consistency Analysis}
\begin{figure}[H]
\centering
\includegraphics[width=0.95\linewidth]{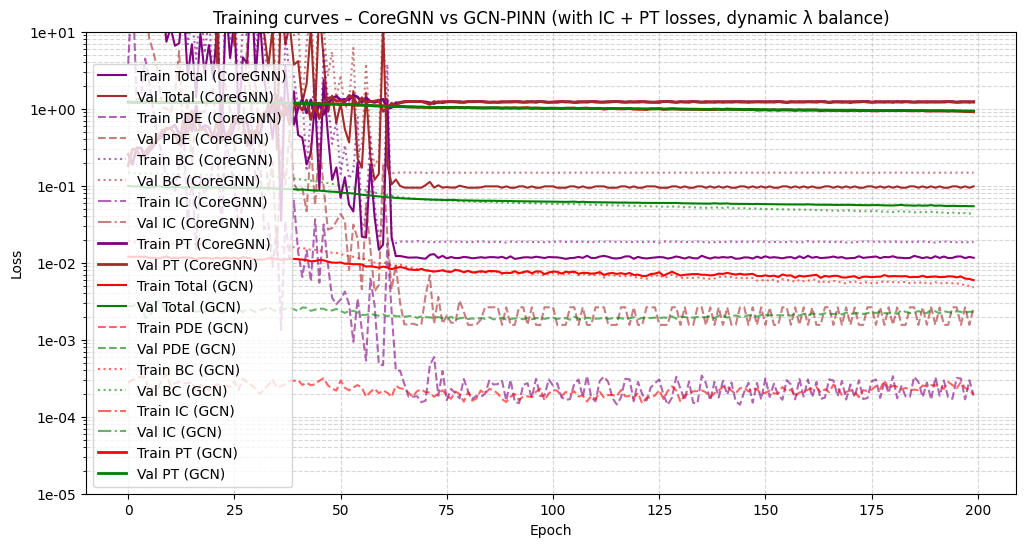}
\caption{
Training and validation loss curves for \textbf{CoreGNN\_Improved} and \textbf{GCN-PINN},
including total, PDE, BC, IC, and P-tensor (PT) losses under dynamic $\lambda$ balancing.
The CoreGNN model exhibits faster convergence, smoother loss decay, and lower steady-state
PDE residuals compared with the GCN baseline, demonstrating improved physical consistency
and numerical stability across training.
}
\label{fig:loss_comparison}
\end{figure}

As shown in Figure~\ref{fig:loss_comparison}, the CoreGNN\_Improved model
converges more slowly and exhibits higher oscillatory frequency compared with
the GCN baseline, but eventually reaches a substantially lower steady-state loss.
This behavior reflects the stronger coupling between node--edge updates in CoreGNN,
which requires iterative adjustment of spatial and temporal consistency.

To interpret this behavior quantitatively, we examined the diagnostic logs at
epoch~200. The CoreGNN model reported
$\dot{f}_{\mathrm{tr}} = 7.58\times10^{-2}$,
$L_{\mathrm{tr}} = 1.52\times10^{-2}$,
and a scaling factor $s_{\mathrm{tr}} = 1.36\times10^{-4}$,
while GCN showed
$\dot{f}_{\mathrm{tr}} = 4.78\times10^{-2}$,
$L_{\mathrm{tr}} = 1.84\times10^{-2}$,
and $s_{\mathrm{tr}} = 3.50\times10^{-1}$.
(The reported losses are dimensionless quantities computed on normalized node values) Although CoreGNN displays a smaller scale and a larger raw temporal derivative,
the two terms become comparable after scaling, indicating that the model
has internally adjusted the ratio between temporal and spatial components
to enforce the diffusion relation $\dot{f} \approx D L f$.
In contrast, GCN maintains a smoother loss curve but a larger imbalance
between $\dot{f}$ and $L f$, suggesting weaker physical regularization.

Overall, these modifications from CoreGNN (v1.0) to CoreGNN\_Improved (v2.0)
mainly served to stabilize training and make the physical residuals more interpretable.
The addition of anchor correction and dynamic weighting reduced sensitivity
to boundary errors, while the P-tensor consistency term explicitly linked
node- and edge-level predictions through the discrete incidence matrix.
Time encoding and synchronized rollout updates made the model’s temporal outputs
better aligned with the PDE integration steps. 
Although each change is small, their combination produced a more reliable
and numerically consistent CoreGNN variant, which we use for all subsequent analyses.

\subsection{Results on Real-World Irregular Meshes}
\label{subsec:realworld_results}

\subsubsection{Comparison 3: Stanford Bunny Benchmark}

We next evaluate the proposed \textbf{CoreGNN} model on the
\textbf{Stanford Bunny}, a widely used real-world mesh containing strong curvature
variation, anisotropic sampling neighborhoods, and narrow geometric folds.
This benchmark tests whether the model remains stable under highly irregular
surface geometry beyond synthetically constructed domains.

\begin{table}[H]
\centering
\caption{Stanford Bunny benchmark: comparison of CoreGNN and GCN against CN baselines under normalized and PDE-scaled evaluation metrics.}
\label{tab:bunny}
\begin{tabular}{lcccc}
\toprule
\textbf{Metric Type} & \textbf{Model} & \textbf{MAE} & \textbf{MSE} & \textbf{L$_2$ norm} \\
\midrule
CN\_norm (normalized) &
CoreGNN & $\bm{5.4211\times10^{-3}}$ & $\bm{2.9475\times10^{-5}}$ & $\bm{5.4291\times10^{-3}}$ \\
CN\_norm (normalized) &
GCN-PINN & $7.0816\times10^{-3}$ & $6.8853\times10^{-5}$ & $8.2978\times10^{-3}$ \\
CN\_PDE ($1/d^2$ weight) &
CoreGNN & $3.4686\times10^{-2}$ & $2.1307\times10^{-3}$ & --- \\
CN\_PDE ($1/d^2$ weight) &
GCN-PINN & $\bm{3.4311\times10^{-2}}$ & $\bm{2.0299\times10^{-3}}$ & --- \\
\midrule
\textbf{PDE Residual $L_2$} &
CoreGNN & \multicolumn{3}{c}{$\bm{5.4286\times10^{-3}}$} \\
\textbf{PDE Residual $L_2$} &
GCN-PINN & \multicolumn{3}{c}{$9.6463\times10^{-3}$} \\
\bottomrule
\end{tabular}
\end{table}

On the Bunny surface, CoreGNN\_Improved achieves consistently lower errors under
the CN-normalized metric, while GCN obtains slightly smaller MAE and MSE under the
PDE-scaled $1/d^2$ Laplacian. However, CoreGNN\_Improved attains a substantially
lower PDE residual norm, indicating stronger enforcement of the diffusion relation
on curved real-world geometry.

\subsubsection{Comparison 4: Coconut Inner-Surface Benchmark}

We further evaluate the proposed \textbf{CoreGNN} model on a scanned
\textbf{coconut inner-shell surface}, whose geometry exhibits pronounced local curvature,
highly nonuniform sampling density, and multiple open boundary loops.
This geometry reflects irregularity that naturally arises during physical shell formation
and differs from the synthetic perturbations used in earlier benchmarks.

\begin{table}[H]
\centering
\caption{Coconut inner-surface benchmark: comparison of CoreGNN and GCN against CN baselines
under normalized and PDE-scaled evaluation metrics.}
\label{tab:comp4_coconut}
\begin{tabular}{lcccc}
\toprule
\textbf{Metric Type} & \textbf{Model} & \textbf{MAE} & \textbf{MSE} & \textbf{L$_2$ norm} \\
\midrule
CN\_norm (normalized) &
\textbf{CoreGNN} &
$\bm{7.3925\times10^{-3}}$ &
$\bm{5.7742\times10^{-5}}$ &
$\bm{7.5988\times10^{-3}}$ \\
CN\_norm (normalized) &
GCN &
$7.9178\times10^{-3}$ &
$7.3866\times10^{-5}$ &
$8.5945\times10^{-3}$ \\
\midrule
CN\_PDE ($1/d^2$ weight) &
CoreGNN &
$8.2586\times10^{-3}$ &
$7.4801\times10^{-5}$ &
--- \\
CN\_PDE ($1/d^2$ weight) &
\textbf{GCN} &
$\bm{7.1010\times10^{-3}}$ &
$\bm{5.6560\times10^{-5}}$ &
--- \\
\midrule
\textbf{PDE Residual $L_2$} &
\textbf{CoreGNN} &
\multicolumn{3}{c}{$\bm{7.7321\times10^{-3}}$} \\
\textbf{PDE Residual $L_2$} &
GCN &
\multicolumn{3}{c}{$1.0896\times10^{-2}$} \\
\bottomrule
\end{tabular}
\end{table}

The results show that \textbf{CoreGNN} achieves consistently lower errors than the GCN baseline under the standard CN\_norm evaluation, with a notably smaller PDE residual that reflects stronger enforcement of physical consistency. Under the stricter CN\_PDE weighting, the two models display mixed pointwise accuracy, but CoreGNN maintains a substantially lower residual norm, indicating improved numerical stability under geometry-sensitive conditions. Overall, the gains remain consistent across both evaluation settings, confirming that the enhanced model provides more reliable and physically coherent diffusion predictions.

\subsubsection{Comparison 5: Leaf Surface Benchmark}

Finally, we evaluate the models on a scanned \textbf{leaf surface}, characterized by
thin-sheet geometry, strong anisotropy, and sharp curvature ridges. This mesh
contains the smallest number of vertices among the three real-world surfaces,
resulting in sparse and highly irregular neighborhoods.

\begin{table}[H]
\centering
\caption{Leaf surface benchmark: comparison of CoreGNN and GCN against CN baselines under normalized and PDE-scaled evaluation metrics.}
\label{tab:leaf}
\begin{tabular}{lcccc}
\toprule
\textbf{Metric Type} & \textbf{Model} & \textbf{MAE} & \textbf{MSE} & \textbf{L$_2$ norm} \\
\midrule
CN\_norm (normalized) &
CoreGNN & $\bm{1.6781\times10^{-3}}$ &
$\bm{4.0948\times10^{-6}}$ & $\bm{2.0236\times10^{-3}}$ \\
CN\_norm (normalized) &
GCN-PINN & $7.3119\times10^{-3}$ & $1.1372\times10^{-4}$ &
$1.0664\times10^{-2}$ \\
CN\_PDE ($1/d^2$ weight) &
CoreGNN & $\bm{1.2425\times10^{-1}}$ & $\bm{2.9521\times10^{-2}}$ & --- \\
CN\_PDE ($1/d^2$ weight) &
GCN-PINN & $1.2699\times10^{-1}$ & $2.9363\times10^{-2}$ & --- \\
\midrule
\textbf{PDE Residual $L_2$} &
CoreGNN & \multicolumn{3}{c}{$\bm{2.0355\times10^{-3}}$} \\
\textbf{PDE Residual $L_2$} &
GCN-PINN & \multicolumn{3}{c}{$1.3135\times10^{-2}$} \\
\bottomrule
\end{tabular}
\end{table}

On the leaf surface, \textbf{CoreGNN\_Improved} shows the clearest advantage among the
real-world meshes, achieving markedly lower errors than GCN under the CN\_norm
evaluation across MAE, MSE, and L$_2$ metrics. Under the stricter CN\_PDE weighting,
the two models exhibit mixed pointwise accuracy, but \textbf{CoreGNN\_Improved}
maintains a lower PDE residual norm, indicating stronger physical consistency on this
highly anisotropic thin-shell geometry.

\subsection{Discussions}
\label{sec:discussion}

The experimental results demonstrate that the proposed CoreGNN\_Improved model consistently 
achieves stable and physically coherent diffusion predictions across a broad spectrum of 
irregular geometries. Although diffusion is used as the primary evaluation task, the 
performance gains arise from the structural design of the model rather than the specific 
equation being solved. The node–edge–edge–node propagation mechanism enables the network to 
represent local gradients and flux interactions directly on the graph, while the 
P-tensor consistency loss enforces a coherent coupling between node states and edge-level 
quantities. This operator-consistent structure provides a principled way to integrate 
discrete differential operators into neural message passing.

Across various mesh categories—including randomly sampled surfaces, physically constructed 
irregular surfaces, and real-world scanned geometries—the model consistently achieves 
lower prediction errors and smaller PDE residuals than standard graph-convolutional 
baselines.
These results highlight the model’s 
ability to generalize across strong curvature variation, anisotropic sampling, and complex 
open boundaries, all of which pose significant challenges for both numerical solvers and 
learning-based approaches.

Overall, these findings suggest that embedding operator-level structure into GNNs is 
more effective than simply enlarging model capacity or relying on data-driven optimization 
alone. By preserving the discrete relationship between gradient and divergence operators, 
CoreGNN\_Improved produces dynamics that remain stable, interpretable, and robust on 
irregular meshes. This framework thus offers a practical pathway for bridging numerical PDE 
principles with modern graph-based learning, enabling reliable scientific computation on 
complex geometric domains.

\section{Future Work}

\subsection{Broader Construction of Irregular-Mesh Datasets}

Future work will explore richer strategies for constructing irregular meshes beyond point-cloud
sampling and KNN-based connectivity. This includes classical geometric methods such as Delaunay
triangulation and surface-based reconstruction techniques. In addition, 3D geometry processing
toolkits such as Open3D—which provide Poisson reconstruction, mesh simplification, resampling, and
normal estimation—will be used to improve the surface-to-graph conversion pipeline. The objective is
to better preserve curvature, local topology, and neighborhood structure when converting raw meshes
into graph representations.

\subsection{Extension to More Complex PDE Models}

With the spatial and temporal components of the model now fully implemented, an important next
step is to extend the framework to more complex PDEs, including nonlinear reaction–diffusion systems
and other coupled multi-field equations. These settings will impose stronger physical constraints on the
learned operators and provide a more stringent evaluation of operator-level consistency on irregular
geometries.

\subsection{Comparison with Additional Numerical and Learning-Based Solvers}

Since the current model supports time-dependent rollouts, future evaluations will compare predictions
at fixed time points against a broader family of baselines, including classical numerical schemes,
graph-based PDE solvers, operator-learning models, and PINN variants. Performing these comparisons
on the same irregular meshes will provide a more complete assessment of the strengths and limitations
of the proposed OCGNN-PINN framework.

\section*{Acknowledgments}
Thank you to Risi Kondor for his advice, suggestions, and feedback on this work.


\newpage
\begin{thebibliography}{99}

\bibitem{CFL_source_citation}
Courant, R., Friedrichs, K., and Lewy, H. (1928). 
\textit{Über die partiellen Differenzengleichungen der mathematischen Physik}. 
\textbf{Mathematische Annalen}, 100(1), 32–74. 
(English translation: *On the Partial Difference Equations of Mathematical Physics*, IBM Journal, 1967.)

\bibitem{chung1997spectral}
Chung, F. R. K. (1997).
\textit{Spectral Graph Theory}.
American Mathematical Society.

\bibitem{kondor2018generalization}
Kondor, R., and Trivedi, S. (2018).
``On the generalization of equivariance and convolution in neural networks,''
\textit{Advances in Neural Information Processing Systems (NeurIPS)},2018.

\bibitem{raissi2019physics}
M.~Raissi, P.~Perdikaris, and G.~E. Karniadakis, 
``Physics-informed neural networks: A deep learning framework for solving forward and inverse problems involving nonlinear partial differential equations,'' 
\textit{Journal of Computational Physics}, vol.~378, pp.~686--707, 2019.

\bibitem{battaglia2018relational}
P.~W. Battaglia, J.~B. Hamrick, V.~Bapst, A.~Sanchez-Gonzalez, V.~Zambaldi, M.~Malinowski, A.~Tacchetti, D.~Raposo, A.~Santoro, R.~Faulkner, and others, 
``Relational inductive biases, deep learning, and graph networks,'' 
\textit{arXiv preprint arXiv:1806.01261}, 2018.

\bibitem{crank1975diffusion}
J.~Crank, 
\textit{The Mathematics of Diffusion}, 
2nd~ed., Oxford University Press, 1975.

\bibitem{sundar2025sequential}
R.~Sundar et al., 
``Sequential learning based PINNs to overcome temporal domain complexities in unsteady flow past flapping wings,'' 
ScienceDirect, 2025.

\bibitem{leveque2007finite}
R.~J. LeVeque,
\textit{Finite Difference Methods for Ordinary and Partial Differential Equations: 
Steady-State and Time-Dependent Problems}.
SIAM, 2007.

\bibitem{kipf2017semi}
T.~N. Kipf and M.~Welling,
``Semi-Supervised Classification with Graph Convolutional Networks,''
\textit{International Conference on Learning Representations (ICLR)}, 2017.

\bibitem{oono2020graph}
K.~Oono and T.~Suzuki,
``Graph Neural Networks Exponentially Lose Expressive Power for Node Classification,''
\textit{International Conference on Learning Representations (ICLR)}, 2020.

\bibitem{battaglia2018relational}
P.~W. Battaglia, J.~B. Hamrick, et al.,
``Relational inductive biases, deep learning, and graph networks,''
\textit{Nature}, pp.~428--436, 2018.

\bibitem{trask2022ddec}
N.~Trask, A.~Huang, and X.~Hu,
``Enforcing exact physics in scientific machine learning: A data-driven exterior calculus on graphs,''
\textit{Journal of Computational Physics}, vol.~456, pp.~110969, 2022.
```

\bibitem{pfaff2021learning}
T.~Pfaff, A.~Sanchez-Gonzalez, S.~M. Cranmer, and P.~Battaglia,
``Learning Mesh-Based Simulation with Graph Networks,''
\textit{International Conference on Learning Representations (ICLR)}, 2021.

\bibitem{brandstetter2022message}
J.~Brandstetter, D.~E. Worrall, and M.~Welling,
``Message Passing Neural PDE Solvers,''
\textit{Advances in Neural Information Processing Systems (NeurIPS)}, 2022.

\bibitem{bridson2007fast}
Bridson, R. (2007).
Fast Poisson disk sampling in arbitrary dimensions.
\textit{ACM SIGGRAPH 2007 Sketches}, 22(1), 1–1.
doi:10.1145/1278780.1278807

\bibitem{enakoutsa2017physics}
K. Enakoutsa, Y. Hammi, J. E. Crawford, “Physics Based Damage Model of Composites for High Speed Structures,” presented to MREC\(^3\), March 18, 2017. \\
DOI: 10.13140/RG.2.2.24742.78408.  
Available at: \url{https://www.researchgate.net/publication/329209586_CCResearch_Physics_Based_Damage_Model}

\bibitem{kutuk2024allen}
K.~Kütük and E.~Yücel,
\newblock ``Energy dissipation preserving physics-informed neural networks for
Allen--Cahn equations,''
\newblock {\em arXiv preprint arXiv:2404.12384}, 2024.


\bibitem{Guo2017CNStability}
Guo, B., Xu, M., \& Zhang, F. (2017).
\newblock Stability and convergence of the Crank--Nicolson scheme for a class of variable-coefficient tempered fractional diffusion equations.
\newblock \emph{Advances in Continuous and Discrete Models}, \textbf{2017}(1), 1150.
\newblock Springer. DOI:10.1186/s13662-017-1150-1.

\bibitem{Hein2007Laplacian}
Hein, M., Audibert, J.-Y., \& von Luxburg, U. (2007).
\newblock Graph Laplacians and their convergence on random neighborhood graphs.
\newblock \emph{Journal of Machine Learning Research}, \textbf{8}, 1325--1368.

\bibitem{Hands2024Ptensors}
Hands, A. R., Sun, T., \& Kondor, R. (2024).
\newblock P-tensors: A general framework for higher order message passing in subgraph neural networks.
\newblock \emph{Proceedings of the 27th International Conference on Artificial Intelligence and Statistics}, \textbf{238}, 424--432.


\end{thebibliography}
\end{document}